\documentclass[sigconf, screen]{acmart}
\AtBeginDocument{%
  }

\setcopyright{acmlicensed}
\copyrightyear{2025}
\acmYear{2025}
\acmDOI{XXXXXXX.XXXXXXX}
\acmConference[Conference ACM MultiMedia]{Make sure to enter the correct
  conference title from your rights confirmation email}{October 27--31,
  2025}{Dublin, Ireland}
\acmISBN{978-1-4503-XXXX-X/2018/06}

\begin{document}
\title{
  "Hi AirStar, Guide Me to the Badminton Court."
  }

\author{Ziqin Wang}
\email{wzqin@buaa.edu.cn}
\author{Jinyu Chen}
\email{chenjinyu@buaa.edu.cn}
\affiliation{%
  \institution{Beihang University}
  \city{Haidian}
  \state{Beijing}
  \country{China}
}

\author{Xiangyi Zheng}
\email{21241102@buaa.edu.cn}
\author{Qinan Liao}
\email{21241019@buaa.edu.cn }
\affiliation{%
  \institution{Beihang University}
  \city{Haidian}
  \state{Beijing}
  \country{China}
}

\author{Linjiang Huang}
\email{ljhuang@buaa.edu.cn}
\author{Si Liu}
\email{liusi@buaa.edu.cn}
\affiliation{%
  \institution{Beihang University}
  \city{Haidian}
  \state{Beijing}
  \country{China}
}

\renewcommand{\shortauthors}{Wang et al.}

\begin{abstract}

Unmanned Aerial Vehicles, operating in environments with relatively few obstacles, offer high maneuverability and full three-dimensional mobility. 
This allows them to rapidly approach objects and perform a wide range of tasks often challenging for ground robots, making them ideal for exploration, inspection, aerial imaging, and everyday assistance. 
In this paper, we introduce AirStar, a UAV-centric embodied platform that turns a UAV into an intelligent aerial assistant: a large language model acts as the cognitive core for environmental understanding, contextual reasoning, and task planning. 
AirStar accepts natural interaction through voice commands and gestures, removing the need for a remote controller and significantly broadening its user base.
It combines geospatial knowledge-driven long-distance navigation with contextual reasoning for fine-grained short-range control, resulting in an efficient and accurate vision-and-language navigation (VLN) capability.
Furthermore, the system also offers built-in capabilities such as cross-modal question answering, intelligent filming, and target tracking.
With a highly extensible framework, it supports seamless integration of new functionalities, paving the way toward a general-purpose, instruction-driven intelligent UAV agent.
The supplementary PPT is available at
\href{https://buaa-colalab.github.io/airstar.github.io}{https://buaa-colalab.github.io/airstar.github.io}.

\end{abstract}

\ccsdesc[500]{Computing methodologies~Planning and scheduling}
\ccsdesc[500]{Computing methodologies~Computer vision}

\keywords{Unmanned Aerial Vehicles, Intelligent Agent, Large Language Model, Vision and Language Navigation}

\begin{teaserfigure}
\centering
\includegraphics[width=0.98\textwidth]{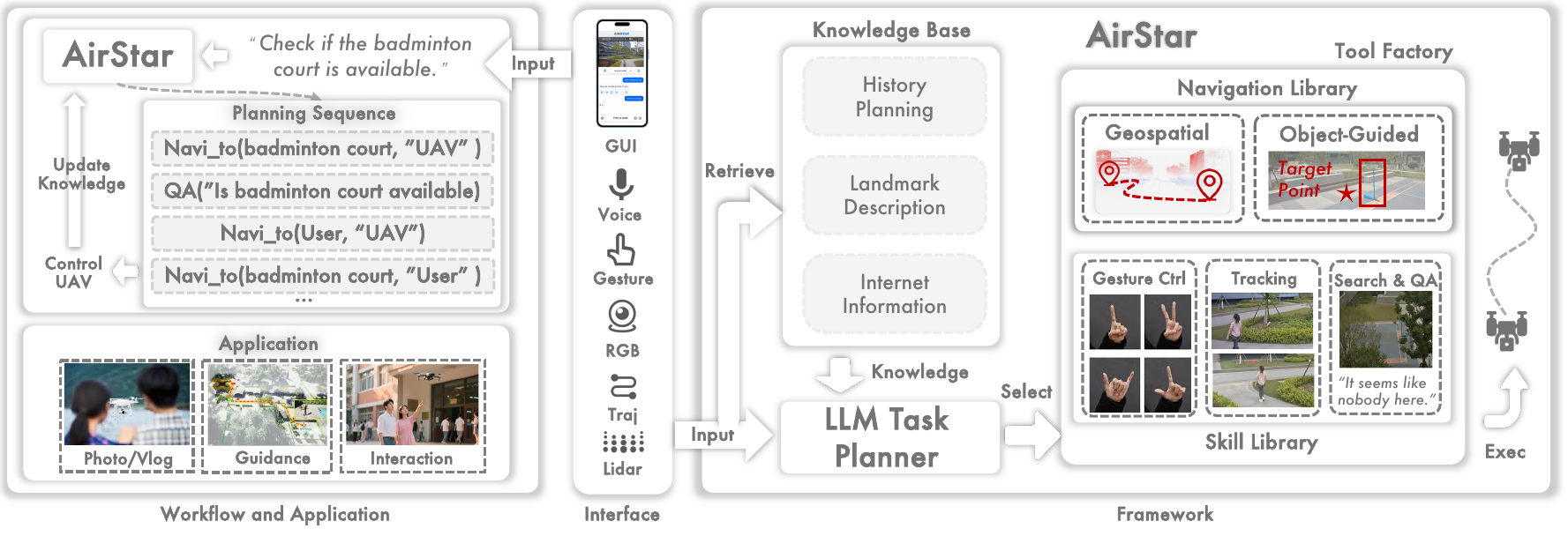}
  \caption{Overall of the AirStar. The left diagram depicts an example of a task reasoning workflow and application scenarios. The diagram to the right details the system’s framework. AirStar mainly contains three parts, namely the LLM Task Planner, the Knowledge Base, and the Tool Factory. By decomposing the task into an API list, AirStar can efficiently execute numerous complex tasks.
}
\vspace{3mm}
  \label{fig:teaser}
\end{teaserfigure}

\settopmatter{printacmref=false}
\renewcommand\footnotetextcopyrightpermission[1]{}

\received{20 February 2007}
\received[revised]{12 March 2009}
\received[accepted]{5 June 2009}

\maketitle

\section{Introduction}
Unmanned Aerial Vehicles (UAVs) have demonstrated significant potential as advanced aerial mobile platforms, widely utilized in environmental detection and aerial photography.
Their flexible maneuverability, multi-perspective observation, adaptability to complex terrains, and three-dimensional mobility give them distinct advantages over ground robots, making them indispensable in unmanned systems.
However, the majority of current UAVs remain heavily dependent on manual operation and exhibit limited intelligence. Even with some autonomous features, their capabilities are typically restricted to isolated tasks such as detection~\cite{fang2023danet,feng2024u2udata} or navigation~\cite{yin2024autonomous}, falling short of achieving full-scenario intelligence that enables comprehensive perception, understanding, and decision-making across diverse and dynamic environments.
%

In this paper, we propose an innovative system, \textbf{AirStar}, which seamlessly integrates UAVs’ mobility with Large Language Model (LLM)-driven decision-making to unlock new possibilities in autonomous aerial systems.
%
%
%
%
AirStar leverages an LLM-based agent~\cite{huang2024understanding} as a high-level task planner. By analyzing mission instructions, historical data, and environmental priors, the agent iteratively decomposes complex tasks, selects and composes suitable skills from the tool factory, and executes these skills to control the UAV, enabling comprehensive, scenario-aware decision-making and adaptation.

The tool factory includes two core components: 
1) Navigation Library, which contains: a modular Geospatial Aware Navigation Module for long-range navigation based on off-line map modeling and a flexible Object-Guided Navigation Module for short-range navigation based on visual-language understanding.
2) Skill library, which mainly includes implementations of interactive functions, such as target tracking, gesture recognition, and intelligent question answering.
%
%
With dynamic environmental perception and various interaction methods, AirStar supports advanced applications such as hands-free, automatic following for photography and vlogging, navigation guidance, and intelligent interactions.

\section{System Framework}
\subsection{LLM Task Planner}
To enhance the efficiency of model processing for user instructions and mitigate hallucinations, the framework first jointly queries a knowledge base containing historical plans/navigation records, landmark descriptions, and up-to-date internet information, using both the user's instruction and the UAV's perception inputs. 
The retrieved knowledge is combined with the user's instruction and information about available tool functions, including their required parameters and usage descriptions from the tool library, to form the input for the LLM agent.
The agent decomposes complex tasks through chain-of-thought reasoning, maps each step to executable skills along with corresponding parameters, and yields a structured task planning. 
The UAV executes the resulting sequence of skills to accomplish the user-provided instruction.
Each executable skill is equipped with a mechanism to detect execution success or failure. Upon failure, the LLM task planner initiates replanning informed by the execution history and the underlying cause of the failure.

\subsection{Navigation Library}
AirStar’s capability for language-guided navigation serves as the foundational component that empowers various applications. To support diverse navigation tasks, it is designed to handle two primary scenarios: long-range navigation that leverages geographical knowledge (\textit{e.g.}, ``Go to Central Campus 
Street'') and short-range navigation that relies on visual comprehension (\textit{e.g.}, ``Fly ahead of the tree''). To this end, we introduce two specialized modules: Geospatial-Aware Navigation and Object-Guided Navigation.  

\vspace{1mm}
\noindent\textbf{Geospatial Aware Navigation.} 
For long-range navigation tasks that rely on landmark information, we construct a discrete, node-based geospatial landmark map. Each node within this map comprises precise GPS coordinates accompanied by annotations denoting landmark orientations (\textit{e.g.}, Teaching Building North).
Additionally, we design two occupancy maps tailored respectively for UAV autonomous exploration and pedestrian guidance, enabling AirStar to select optimal routes according to mission requirements.

During navigation, AirStar initially extracts landmark entities from the provided instructions and retrieves the corresponding GPS coordinates from the landmark map. Subsequently, the A* algorithm computes critical waypoints within the relevant occupancy map. Finally, we incorporate the Ego-Planner~\cite{zhou2020ego} to formulate smooth and feasible trajectories, facilitating reliable long-distance navigation.

\noindent\textbf{Object-Guided  Navigation.} 
%
For navigation tasks within the visual range, we use Qwen2.5-VL~\cite{bai2025qwen2} to predict the target points that the UAV must reach to complete tasks based on instructions.
Since these predicted targets are 2D coordinates, we convert them into corresponding 3D world positions by leveraging depth maps (from the RGB-D camera) and the camera's intrinsic/extrinsic parameters. AirStar then employs Ego-Planner to navigate precisely toward these mapped 3D coordinates. 
AirStar then employs Ego-Planner to navigate precisely toward these mapped 3D coordinates. This two-stage workflow enables accurate short-range navigation based on visual-semantic reasoning.

\vspace{-2mm}
\subsection{Skill Library}
To enhance AirStar's capabilities, we have designed a Skill Library that integrates advanced UAV functionalities, including gesture control, tracking, and search \& QA. The LLM Task Planner can access these skills to handle complex requirements.

\vspace{1mm}
\noindent\textbf{Gesture Control} This module uses an on-board human detection algorithm to initialize camera framing by locating subjects. Users can further refine viewpoints with gesture-based commands (up/down/left/right/forward/backward) for fine-grained control.

\vspace{1mm}
\noindent\textbf{Tracking.} 
AirStar supports target initialization via either instructions (via GroundingDINO~\cite{liu2024grounding}) or interactive clicking in the feedback interface, followed by real-time target tracking through LightTrack~\cite{yan2021lighttrack}. The UAV adjusts its position to keep the target in the view center while avoiding occlusions based on lidar observation.

\vspace{1mm}
\noindent\textbf{Search \& QA.} 
After the UAV navigates to the question-relevant area, the candidate viewing angle is computed based on the GPS coordinates of the relevant landmark and the UAV's current position. To account for potential GPS drift, we scan adjacent viewing angles and evaluate their SigLIP similarity scores with the landmark-related nouns. The view with the highest score is selected for VLM inference to generate the answer.

\section{ System Workflow and Infrastructure}
%
%
%
The system primarily enables user interaction through a smartphone application, supporting multiple input modalities including text and voice commands. Upon activation, the UAV automatically ascends to a hovering position and enters standby mode. When navigation instructions are received, the UAV executes a sequence of API calls orchestrated by AirStar to accomplish the specified tasks. After task completion, the UAV autonomously returns near the user and resumes standby hovering.

Furthermore, the UAV is equipped with a network card enabling Wi-Fi communication with the smartphone and 5G connectivity with the base station. Considering the UAV's limited onboard computational capability, only lightweight algorithms such as obstacle avoidance, target tracking, and gesture recognition run onboard, while computationally intensive LLM-related algorithms are hosted entirely on the base station.


\bibliographystyle{ACM-Reference-Format}
\bibliography{sample-base}

\end{document}